\documentclass[11pt]{article}
\usepackage[super,sort&compress,comma]{natbib}
\usepackage{graphicx}
\graphicspath{{figures/}{./}}
\usepackage{amsmath,amssymb}
\usepackage{booktabs}
\usepackage{xcolor}
\usepackage[margin=1in]{geometry}
\usepackage{hyperref}
\hypersetup{colorlinks,citecolor=blue,linkcolor=blue,urlcolor=blue}
\title{AgentCT: A Vision-Language Agent for Universal CT Imaging}
\author{Zhihao Chen$^{1}$, Qi Gao$^{1}$, Chuang Niu$^{2}$,\\[0.2em]
Qiegen Liu$^{3}$, Yi Zhang$^{4}$, Hongming Shan$^{1}$\\[0.6em]
\small $^{1}$Institute of Science and Technology for Brain-inspired Intelligence,\\
\small Fudan University, Shanghai 200433, China\\
\small $^{2}$School of Instrumentation Science and Engineering,\\
\small Harbin Institute of Technology, Harbin 150001, China\\
\small $^{3}$School of Information Engineering, Nanchang University,\\
\small Nanchang 330031, China\\
\small $^{4}$School of Cyber Science and Engineering, Sichuan University,\\
\small Chengdu 610065, China\\[0.4em]
\small Corresponding author: Hongming Shan (hmshan@fudan.edu.cn)}
\date{}

\begin{document}
\maketitle

\begin{abstract}
Computed tomography (CT) restoration methods often use a single network for low-dose CT (LDCT), sparse-view CT (SVCT), and metal artifact reduction (MAR), despite the distinct physics of these degradations. We propose AgentCT, a vision-language agent that identifies the degradation and routes images to task-specific restoration models. A LoRA-adapted Qwen3-VL-4B controller predicts degradation type and severity, selects a specialist, and reassesses the restored image. The LDCT specialist uses an image-domain diffusion model, whereas the SVCT and MAR specialists combine projection-domain correction, data consistency, differentiable filtered back-projection, and image refinement. We evaluated the controller and specialists separately on patient-disjoint Mayo 2020 abdominal CT data across three severity levels for each degradation. The controller achieved 95.7\% accuracy for degradation type and 93.9\% for severity, with a 100\% JSON parse rate. Using ground-truth degradation labels for routing, AgentCT achieved an overall PSNR/SSIM of 46.18~dB/0.984, compared with 44.93~dB/0.976 for AMIR and 44.29~dB/0.979 for TAMP, and attained the highest PSNR in all nine settings. For SVCT, its dual-domain specialist achieved 46.04~dB with 4.2 million parameters, compared with 44.43~dB for single-task AMIR and 43.57~dB for all-in-one AMIR with 23.5 million parameters. These results support degradation-specific CT restoration with controller-based model selection; evaluation under automatic routing and on independent clinical data remains necessary. Code is available at \url{https://github.com/hao1635/AgentCT}.
\end{abstract}

\vspace{2mm}
\noindent{\bf Keywords:} Universal CT imaging, vision-language model,
agent, low-dose CT, sparse view CT, metal artifact reduction.

\section{Introduction}
\label{sec:intro}
Computed tomography (CT) is a widely used imaging technique for clinical diagnosis and
screening. Because the ionizing radiation of CT carries health risk, the as-low-as-reasonably-achievable principle motivates reducing the radiation dose or the number
of projection views whenever diagnostic image quality can still be maintained~\cite{kiss2026denoising}. In addition, metal implants such as dental fillings, hip prostheses, and spinal fixation corrupt the X-ray measurements. These conditions produce three common CT degradations: low-dose quantum noise
(LDCT), sparse-view streak artifacts (SVCT), and metal artifacts (MAR) arising from beam hardening and photon starvation. All three degrade image quality and may compromise diagnosis and downstream quantitative analysis.

Deep learning has substantially advanced the restoration of individual CT degradations~\cite{barbano2025steerable,kiss2025benchmarking}. For
LDCT denoising, image-domain networks such as RED-CNN~\cite{redcnn} and WGAN-VGG~\cite{wganvgg},
together with recent diffusion models~\cite{corediff,gao2025need}, achieve strong performance. 
For SVCT reconstruction, dual-domain methods such as Learned Primal-Dual~\cite{lpd}, DuDoTrans~\cite{dudotrans}, and FreeSeed~\cite{freeseed} jointly exploit the projection and image
domains to suppress undersampling artifacts. For MAR, DuDoNet~\cite{dudonet} couples the sinogram and image domains, and InDuDoNet~\cite{indudonet} further incorporates projection-trace correction. These degradation-specific models are effective, but each targets
a single task and cannot automatically identify the input degradation or select the matching
model; a human operator must therefore choose the model for each scanning condition.

To alleviate this manual selection burden, universal restoration models handle multiple degradations within a single network. PromptIR~\cite{promptir} encodes degradations with learnable prompts, AMIR~\cite{amir} adaptively routes tasks on a Restormer~\cite{restormer} backbone, and
TAMP~\cite{tamp} is a foundation model for universal enhancement of non-ideal-measurement CT.
Although these approaches avoid manual selection, they rely on a shared architecture across degradations and may overlook their distinct imaging physics. In particular, SVCT and MAR originate from undersampled or corrupted measurements in the sinogram domain, where measurement
consistency can be explicitly enforced~\cite{arridge2026deepprior,hertrich2026regularization}; a shared image-domain network cannot impose this consistency where the corruption actually occurs.

Agents based on large language and vision-language models (VLMs) use model-generated decisions to invoke external tools~\cite{react,fallahpour2025medrax,schmidgall2025agentlaboratory}. In image restoration, recent studies
have applied this approach to identifying degradations and selecting or sequencing restoration models~\cite{agenticir,restoreagent}. These studies motivate investigating whether a vision-language controller can select CT restoration tools that operate
in different domains.

In this work we propose AgentCT, a VLM-based agent for universal CT imaging and, to our knowledge, the first agentic CT framework. AgentCT follows a perceive-route-restore-verify workflow. A single Qwen3-VL-4B controller perceives the degradation type and severity, produces a language diagnosis and a structured routing action, and dispatches the matching physics-aware specialist. LDCT is handled by an image-domain diffusion denoiser, whereas SVCT and MAR are handled by dual-domain networks that restore the sinogram, enforce measurement consistency, and
refine the image after a differentiable filtered back-projection (FBP). After restoration, the controller re-perceives the output to verify whether residual degradation remains, and it declines to restore a clean input. Our contributions are as follows.
\begin{itemize}
\item We develop a vision-language controller for CT degradation recognition, tool selection, and post-restoration assessment.
\item We integrate an image-domain diffusion denoiser for LDCT with dual-domain specialists for SVCT and MAR, allowing the restoration tools to use different
architectures and input domains.
\item With ground-truth routing, the specialists improve overall PSNR by 1.25~dB over AMIR and 1.89~dB over TAMP on nine degradation-severity settings. An SVCT ablation shows that single-task AMIR improves over all-in-one AMIR, while the dual-domain specialist provides a further improvement.
\end{itemize}

\section{Materials and Methods}
\label{sec:methods}

\subsection{Overview}
\label{sec:overview}
\begin{figure}[t]\centering
\includegraphics[width=\textwidth]{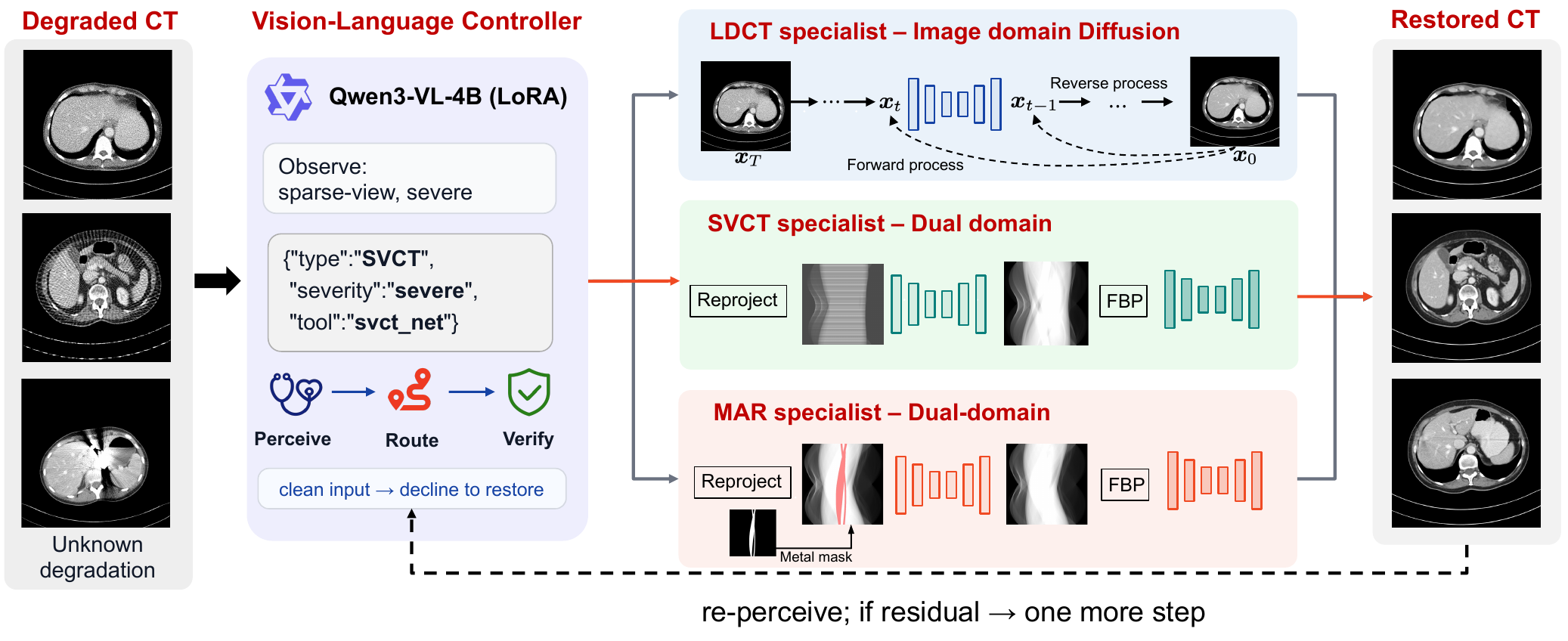}
\caption{\textbf{Overview of AgentCT.} A LoRA-adapted Qwen3-VL-4B controller identifies the degradation type and severity and selects a restoration specialist. LDCT uses an image-domain diffusion denoiser; SVCT and MAR use dual-domain networks. The restored image is returned to the controller for assessment, with at most one additional restoration call. Inputs classified as clean bypass restoration.}
\label{fig:framework}
\end{figure}

As illustrated in Fig.~\ref{fig:framework}, AgentCT consists of a vision-language controller and three restoration specialists. The controller receives a windowed CT image and generates a degradation assessment and a structured tool-selection command.
The selected specialist performs restoration, and its output is returned to the controller for assessment. Inputs classified as clean are returned without restoration.
The specialist inputs and operations are described in Sec.~\ref{sec:specialists}.

\subsection{Vision-language controller}
\label{sec:controller}
We employ Qwen3-VL-4B~\cite{qwenvl} as the controller and adapt it to CT degradation perception
and tool routing using low-rank adaptation (LoRA)~\cite{lora}. Each CT slice is converted to a
windowed image using the HU range $[-350,350]$, chosen to enhance the visibility of degradation
cues, and is paired with an instruction asking the model to identify the degradation type and
severity, briefly describe its observation, and select the corresponding restoration tool.
Because the degradation settings are known during data construction, the type, severity, and
routing labels are directly assigned to each training sample.

The target response has two parts: a one-sentence observation describing the visible
degradation and a structured JSON action containing the fields \texttt{degradation},
\texttt{severity}, and \texttt{tool}. LDCT, SVCT, and MAR are associated with
\texttt{denoise\_net}, \texttt{svct\_net}, and \texttt{mar\_net}, respectively, and a clean
image is assigned \texttt{none}. 
This response format combines a textual description of the observed degradation with a machine-readable command for tool execution.
During fine-tuning, LoRA modules are inserted into the attention and feed-forward projection layers with rank 16,
scaling factor 32, and dropout 0.05; only the assistant response contributes to the
language-modeling loss, and the instruction tokens are masked. The same controller performs
both the initial perception and the subsequent verification, so the restored image can be
re-evaluated without a separate image-quality model.

\subsection{Physics-aware specialists}
\label{sec:specialists}
The tool pool contains three specialists: a diffusion denoiser for LDCT, a sinogram-completion network with image refinement for SVCT, and a metal-trace correction network with image refinement for MAR.

\subsubsection{LDCT: diffusion denoiser}
For LDCT we adopt CoreDiff~\cite{corediff}, a contextual error-modulated diffusion model
developed for low-dose CT denoising. The official pretrained model initializes the network,
which is then fine-tuned jointly on the three dose levels in our training set. The input is
normalized using the HU range $[-1000,2000]$, and the original $T=10$ diffusion process is
retained. Because our data are single 2D slices while CoreDiff uses a three-channel contextual
input, the center slice is replicated across the three input channels during training and
inference. This specialist focuses on removing low-dose noise without modifying the
reconstruction-domain measurements.

\subsubsection{SVCT: dual-domain restoration}
For SVCT we follow the dual-domain strategy of DuDoNet and DuDoTrans~\cite{dudonet,dudotrans}.
Sparse-view measurements are zero-filled onto a 720-view full angular grid and passed to a
sinogram network together with a binary sampling mask. The network first completes the missing projections; an explicit data-consistency step then restores the originally acquired views so that the measured data remain unchanged. The completed sinogram is reconstructed by a
differentiable FBP and refined by an image-domain network, which receives both the completed-sinogram reconstruction and the sparse-view FBP image, thereby combining projection-domain completion with image-domain correction. Training uses an $L_1$-plus-SSIM image loss together
with sinogram-completion and acquired-view consistency losses. The parallel-beam Radon transform and FBP are implemented in PyTorch to enable end-to-end optimization across the two domains.

\subsubsection{MAR: metal-trace guided dual-domain restoration}
For MAR we follow the dual-domain designs of DuDoNet and InDuDoNet~\cite{dudonet,indudonet},
adapting the projection-domain correction to the metal trace. The degraded image is re-projected
into the sinogram domain, and the metal mask is projected with the same Radon transform to
obtain the metal trace. A sinogram network predicts corrected measurements, after which only the
values inside the metal trace are replaced while the measurements outside the trace are
preserved. The corrected sinogram is reconstructed through differentiable FBP and refined by an
image-domain network using the original artifact image as additional guidance. The network is
optimized with image-domain $L_1$ and SSIM losses together with sinogram supervision. During
training, the known metal region used for MAR synthesis provides the mask; at inference, where
the ground-truth mask is unavailable, the mask is estimated from the degraded HU image by
high-HU thresholding followed by light morphological processing.

\subsection{Routing and verification}
\label{sec:routing}
The structured action generated by the controller connects perception with the specialist pool.
Once a degradation is identified, AgentCT dispatches the corresponding tool and returns its
output to the controller for re-perception; a clean input is assigned no tool and bypasses
restoration. The current implementation allows at most two perception-restoration steps: if the
restored image is verified as clean the process terminates, otherwise one additional tool call
can be issued. Because each test sample here contains a single degradation, one specialist is
generally sufficient; extending the framework to scans with multiple simultaneous degradations
would require studying tool ordering and iterative restoration, which we leave to future work.

\subsection{Datasets and degradation simulation}
\label{sec:data}
Experiments used the publicly available, de-identified Mayo 2020 abdominal CT data from The Cancer Imaging Archive~\cite{moen2021ldct} with a patient-disjoint training--test split. The original data collection was approved by the Mayo Clinic Institutional Review Board, as reported by the dataset authors. This secondary analysis involved no new patient recruitment or access to identifiable patient data. Full-dose CT images served as clean references for restoration and as clean inputs for controller perception and routing.
Nine degradation-severity categories were considered: LDCT at 1/2, 1/4, and 1/10 dose; SVCT with 120, 60, and 30 views; and MAR with titanium (Ti), iron (Fe), and gold (Au) implants. These metal settings were designated
mild, moderate, and severe, respectively. Approximately 15{,}600 degraded--reference image pairs were used for restoration training and 4{,}740 pairs for testing. The controller distinguished the nine degraded categories and the clean category.

For SVCT, a parallel-beam geometry with 720 full views over $180^{\circ}$ was used, from which
120, 60, or 30 views were retained. MAR simulation followed the established polychromatic
protocol~\cite{zhangyu,dudonet}, including a GE 120~kVp spectrum, NIST attenuation coefficients,
water/bone basis decomposition, beam hardening, Poisson noise with an incident photon count of
$N_0 = 2\times10^{7}$, and water-based beam-hardening correction, all reconstructed by FBP. One
hundred metal masks were placed within the patient body region with random spatial
perturbations, and Ti, Fe, and Au represent increasing artifact severity. All images were
normalized using the HU range $[-1000,2000]$ for network training, and quantitative evaluation
used the same window with PSNR and SSIM reported for all methods; qualitative images are
displayed in the abdominal window of $[-160,240]$~HU (WW400/WL40). The reconstruction-domain
specialists were trained on full $512\times512$ images to preserve the global projection
geometry.

\subsection{Baselines and implementation}
\label{sec:impl}
We compared AgentCT with two universal baselines. AMIR~\cite{amir}, a representative all-in-one
medical image restoration model, was retrained on the same LDCT, SVCT, and MAR data. TAMP
~\cite{tamp}, a foundation model for universal enhancement of non-ideal-measurement CT, was
adapted through its intended mechanism: its released MITNet backbone was kept frozen and a
task-specific LoRA adapter (rank 16, scaling 16) was trained on our data for each degradation.
AgentCT uses CoreDiff for LDCT and the dual-domain networks for SVCT and MAR. For a controlled
comparison, each specialist and baseline was evaluated on its matching degradation category so
that restoration quality is measured independently of routing errors. All models were
implemented in PyTorch~2.x with CUDA on NVIDIA GPUs. Each specialist was trained with the AdamW
optimizer and a cosine-decayed learning rate; the dual-domain networks used a combined
$L_1$-plus-SSIM image loss with auxiliary sinogram losses, and the LDCT diffusion denoiser was
warm-started from the public CoreDiff checkpoint and fine-tuned on the three dose levels. The
controller was fine-tuned with LoRA on the perception-routing supervision of
Sec.~\ref{sec:controller}. Hyperparameters and trained weights are released with the code to
support reproducibility.

\subsection{Image-quality evaluation}
\label{sec:metrics}
Restoration quality was evaluated against the full-dose reference using PSNR and SSIM in the $[-1000,2000]$~HU window. RMSE was additionally computed in HU to express the restoration error in physical units. The reported comparisons are descriptive; no formal significance tests were performed.

\begin{table}[h]
\centering
\caption{Perception performance of the vision-language controller on the patient-disjoint test
set.}
\label{tab:perception}
\begin{tabular}{crrr}
\toprule
Method & Type accuracy & Severity accuracy & JSON parse rate \\
\midrule
AgentCT controller & 95.7\% & 93.9\% & 100\% \\
\bottomrule
\end{tabular}
\end{table}

\section{Results}
\label{sec:results}

\begin{table}[t]\centering
\caption{Quantitative comparison on the nine CT degradation categories (PSNR/SSIM), evaluated in
the $[-1000,2000]$~HU window over the full patient-disjoint test set. Best PSNR/SSIM in bold;
$\Delta$ is the PSNR gain of AgentCT over the second-best method in each row.}
\label{tab:main}
\small\begin{tabular}{lccccc}
\toprule
Category & Degraded & AMIR & TAMP & AgentCT (\textbf{ours}) & $\Delta$ \\
\midrule
LDCT 1/2  & 35.84 / .778 & 47.19 / .981 & 47.51 / .989 & \textbf{48.17 / .989} & \textbf{+0.66} \\
LDCT 1/4  & 33.16 / .651 & 46.26 / .978 & 46.41 / .986 & \textbf{47.04 / .987} & \textbf{+0.63} \\
LDCT 1/10 & 28.91 / .441 & 44.70 / .972 & 44.51 / .980 & \textbf{44.97 / .980} & \textbf{+0.27} \\
SVCT 120v & 37.45 / .855 & 48.49 / .989 & 47.64 / .989 & \textbf{50.35 / .993} & \textbf{+1.86} \\
SVCT 60v  & 30.78 / .635 & 43.56 / .965 & 43.41 / .973 & \textbf{46.23 / .985} & \textbf{+2.67} \\
SVCT 30v  & 25.50 / .468 & 38.65 / .943 & 38.90 / .945 & \textbf{41.53 / .967} & \textbf{+2.63} \\
MAR mild (Ti)     & 35.00 / .982 & 50.63 / .996 & 48.58 / .995 & \textbf{51.48 / .996} & \textbf{+0.85} \\
MAR moderate (Fe) & 29.77 / .924 & 46.22 / .991 & 44.45 / .990 & \textbf{46.95 / .992} & \textbf{+0.73} \\
MAR severe (Au)   & 19.92 / .376 & 38.65 / .967 & 37.20 / .962 & \textbf{38.93} / \textbf{.970} & \textbf{+0.28} \\
\midrule
\textbf{Overall} & 30.70 / .679 & 44.93 / .976 & 44.29 / .979 & \textbf{46.18 / .984} & \textbf{+1.25} \\
\bottomrule
\end{tabular}
\end{table}

\subsection{Controller perception and routing}
\label{sec:perc}
To assess whether the controller can reliably perceive CT degradation and generate executable
routing decisions, we evaluated it on a patient-disjoint test set of 2{,}000 images. As shown in
Table~\ref{tab:perception}, the controller achieved 95.7\% accuracy for degradation-type
recognition and 93.9\% for severity estimation, and all generated actions were successfully
parsed into the predefined JSON format. Most recognition errors occurred between mild LDCT and
clean images, where the difference in noise level is visually subtle. These results indicate
that the controller can convert visual perception into structured actions, providing a reliable
interface between CT images and external restoration tools.

\subsection{Quantitative restoration performance}
\label{sec:quant}
Table~\ref{tab:main} reports the per-category restoration performance in the $[-1000,2000]$~HU
window over the full test set. Overall, AgentCT achieved 46.18~dB PSNR and 0.984 SSIM,
outperforming AMIR by 1.25~dB and the TAMP foundation model by 1.89~dB, and it attained the best
PSNR in every category. The improvement was most pronounced for SVCT, where the dual-domain
specialist improved PSNR over the second-best method by 1.86, 2.67, and 2.63~dB at 120, 60, and
30 views, respectively. This benefit arises from explicitly restoring the missing projections
and preserving the acquired measurements through data consistency, which the image-domain AMIR
and TAMP cannot exploit; indeed, the two image-domain baselines performed comparably on these
reconstruction-domain tasks. For MAR, AgentCT achieved PSNR gains of 0.28 to 0.85~dB over the
second-best method while maintaining comparable SSIM. For the three LDCT levels, the diffusion
denoiser improved over the second-best method by 0.27 to 0.66~dB; TAMP was competitive with our
method on LDCT but fell behind on the reconstruction-domain SVCT and MAR tasks, consistent with
its image-domain design.

Table~\ref{tab:hu} reports the error in Hounsfield units. AgentCT achieved the lowest overall
RMSE (17.5~HU), versus 20.2~HU for AMIR and 21.0~HU for TAMP, and it had the lowest RMSE in every
category. The advantage was again largest for the
reconstruction-domain tasks: on 30-view SVCT the RMSE dropped from 34.6-36.0~HU for the
image-domain baselines to 26.0~HU, and on severe metal artifacts (Au) from 37.7-43.7~HU to
35.9~HU, corresponding to smaller HU bias in tissue near the corrupted measurements.

\begin{table}[t]\centering
\caption{Restoration error (RMSE in Hounsfield units), mean~$\pm$~standard deviation over the
full patient-disjoint test set. Best in bold.}
\label{tab:hu}
\begin{tabular}{lccc}
\toprule
Category & AMIR & TAMP & AgentCT (\textbf{ours}) \\
\midrule
LDCT 1/2           & 14.4$\pm$1.3 & 13.2$\pm$0.9 & \textbf{12.8$\pm$1.4} \\
LDCT 1/4           & 15.8$\pm$1.4 & 14.9$\pm$1.2 & \textbf{14.4$\pm$1.5} \\
LDCT 1/10          & 18.6$\pm$1.7 & 18.4$\pm$2.1 & \textbf{17.9$\pm$2.2} \\
SVCT 120v          & 12.4$\pm$1.5 & 13.0$\pm$1.2 & \textbf{10.4$\pm$1.9} \\
SVCT 60v           & 20.9$\pm$2.0 & 20.8$\pm$1.9 & \textbf{15.6$\pm$2.4} \\
SVCT 30v           & 36.0$\pm$5.8 & 34.6$\pm$3.6 & \textbf{26.0$\pm$4.2} \\
MAR mild (Ti)      & 10.2$\pm$2.1 & 11.7$\pm$2.9 & \textbf{9.4$\pm$1.9} \\
MAR moderate (Fe)  & 16.1$\pm$4.8 & 19.0$\pm$6.2 & \textbf{14.8$\pm$4.0} \\
MAR severe (Au)    & 37.7$\pm$13.7 & 43.7$\pm$15.0 & \textbf{35.9$\pm$11.4} \\
\midrule
\textbf{Overall}   & 20.2 & 21.0 & \textbf{17.5} \\
\bottomrule
\end{tabular}
\end{table}

\begin{figure}[ht]\centering
\includegraphics[width=\textwidth]{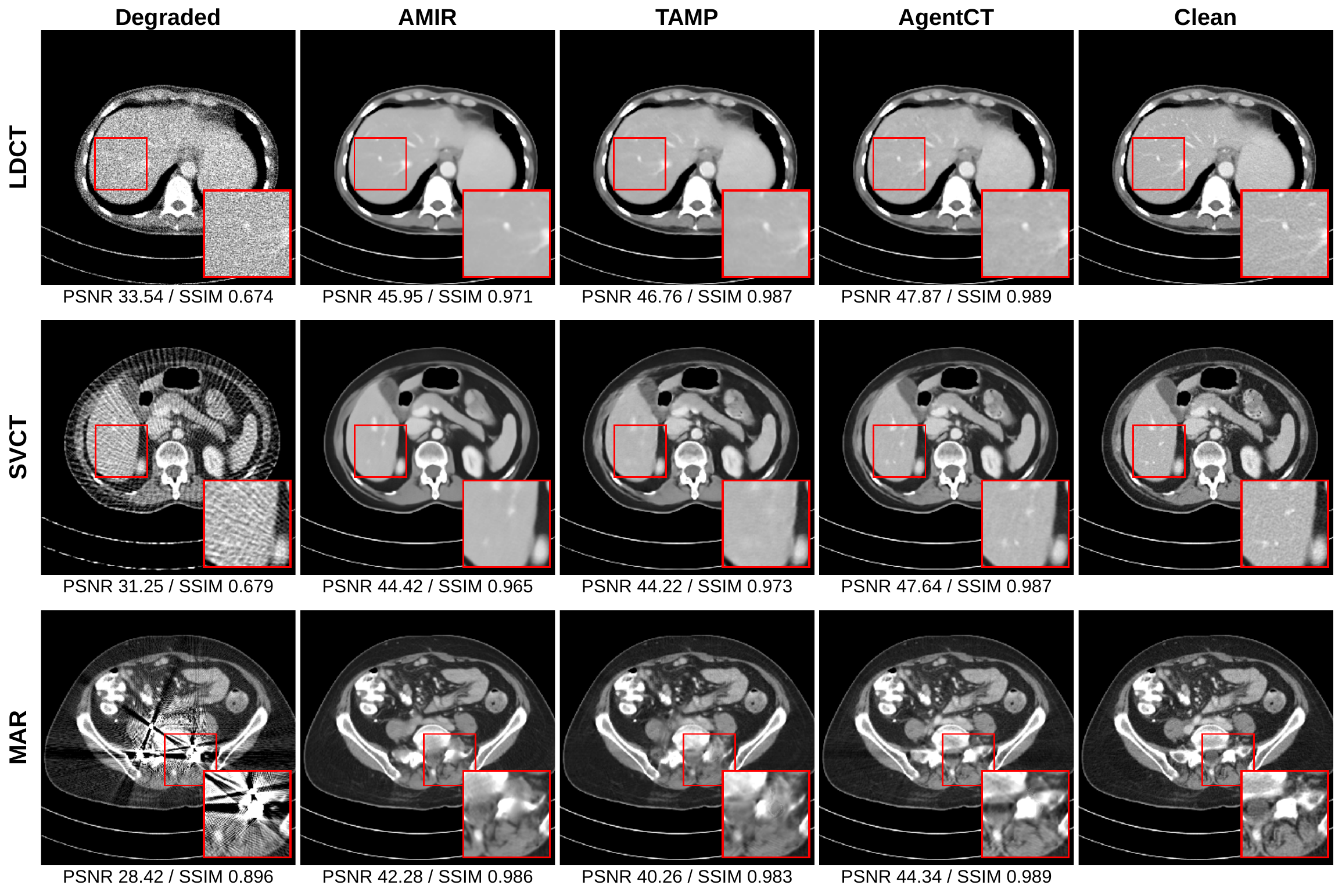}
\caption{Qualitative comparison on LDCT, SVCT, and MAR. Columns show the degraded image, AMIR,
TAMP, AgentCT, and the clean reference. The region of interest is magnified for visual
comparison, and PSNR/SSIM are reported in the $[-1000,2000]$~HU window.}
\label{fig:qual}
\end{figure}

\subsection{Qualitative comparison}
\label{sec:qual}
Fig.~\ref{fig:qual} shows representative restorations with a magnified region of interest for
each degradation. Compared with the AMIR and TAMP baselines, the LDCT specialist suppresses
noise while retaining soft-tissue structure; for SVCT, the dual-domain network removes the
radial streaks of sparse-view acquisition and recovers clearer anatomical boundaries; and for
MAR, the AgentCT specialist suppresses metal-induced streaks and dark bands while preserving the
surrounding anatomy. These observations are consistent with the quantitative results in
Table~\ref{tab:main}.

\begin{figure}[t]\centering
\includegraphics[width=\textwidth]{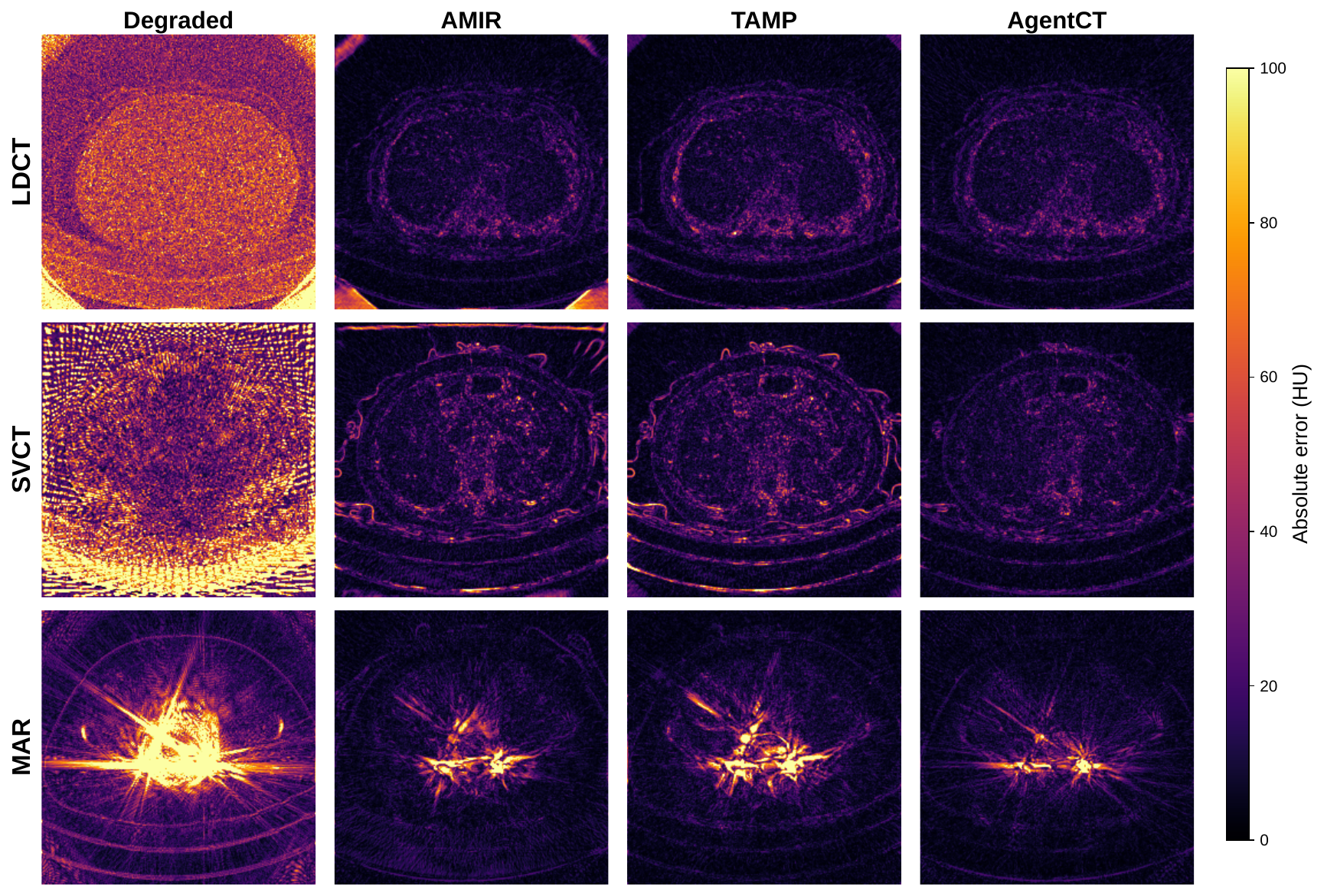}
\caption{Absolute error maps $|\hat{x}-x|$ in Hounsfield units for the degraded input, AMIR,
TAMP, and AgentCT on the three degradations (rows: LDCT, SVCT, MAR), sharing the color scale on
the right. Darker indicates smaller error.}
\label{fig:diff}
\end{figure}

The absolute-error maps in Fig.~\ref{fig:diff} make the differences explicit in Hounsfield units.
AgentCT shows the smallest and most spatially uniform residual error for all three degradations;
the image-domain baselines leave pronounced residual error along the sparse-view streaks and in
the tissue immediately surrounding the metal, precisely the reconstruction-domain regions where
enforcing measurement consistency is most beneficial.

\begin{figure}[!h]\centering
\includegraphics[width=0.7\textwidth]{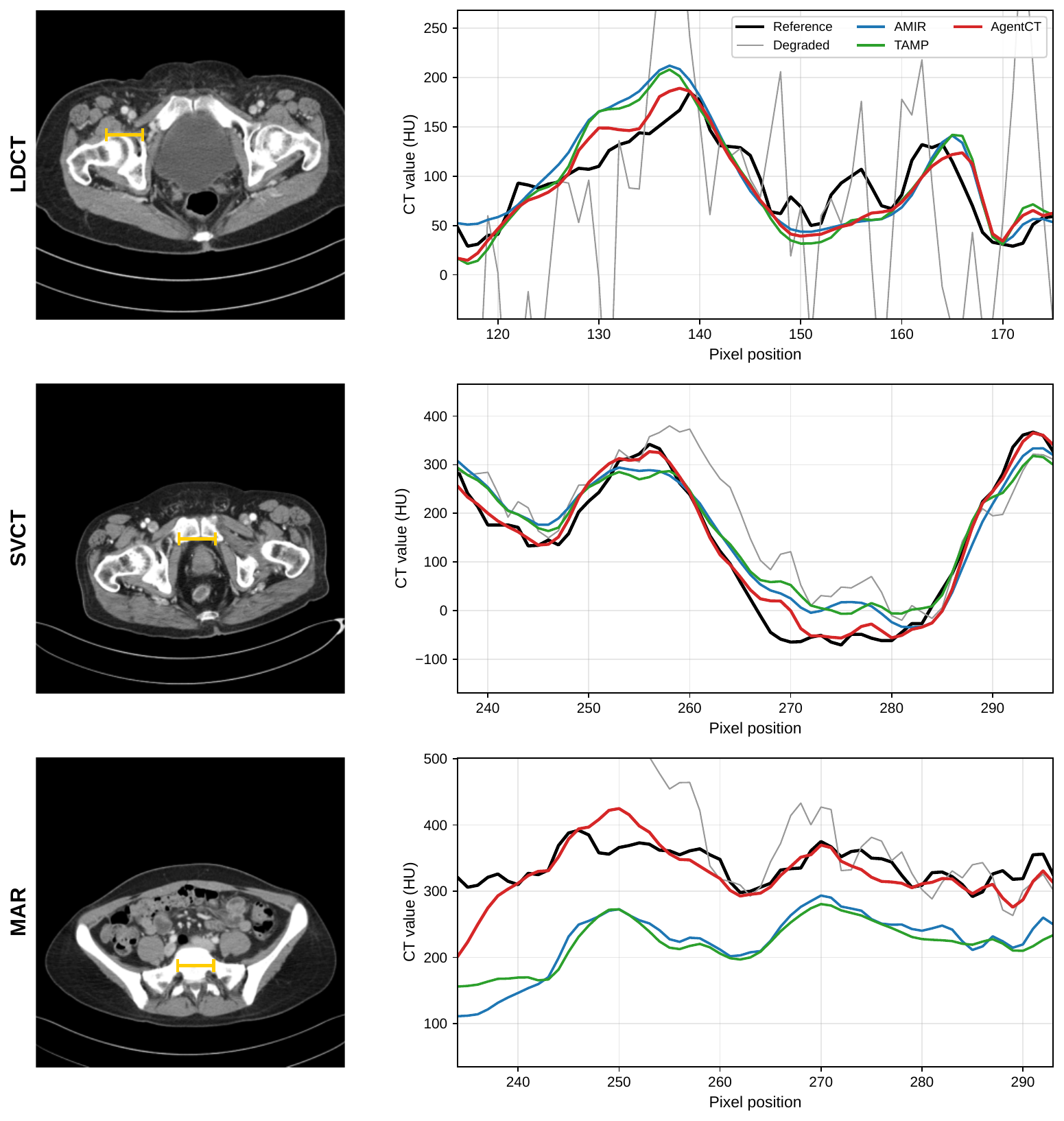}
\caption{CT-value profiles in Hounsfield units along the yellow line segments marked on the reference images (rows: LDCT, SVCT, MAR). Curves represent the degraded input (gray), AMIR (blue), TAMP (green), AgentCT (red), and full-dose reference (black).}
\label{fig:profiles}
\end{figure}

Finally, the consistency of the restored CT values with the reference is examined with short
Hounsfield-unit profiles through structured soft tissue for each degradation
(Fig.~\ref{fig:profiles}). The AgentCT curve (red) closely follows the full-dose reference
(black), whereas the AMIR and TAMP baselines deviate markedly at the same locations---most
evidently for SVCT and MAR, where the image-domain baselines undershoot the soft-tissue
attenuation by roughly 25 and 100~HU, respectively, that the measurement-consistent dual-domain
specialist recovers. This confirms that AgentCT restores quantitatively accurate CT numbers
rather than merely a visually plausible image.

\begin{figure}[!t]\centering
\includegraphics[width=0.9\textwidth]{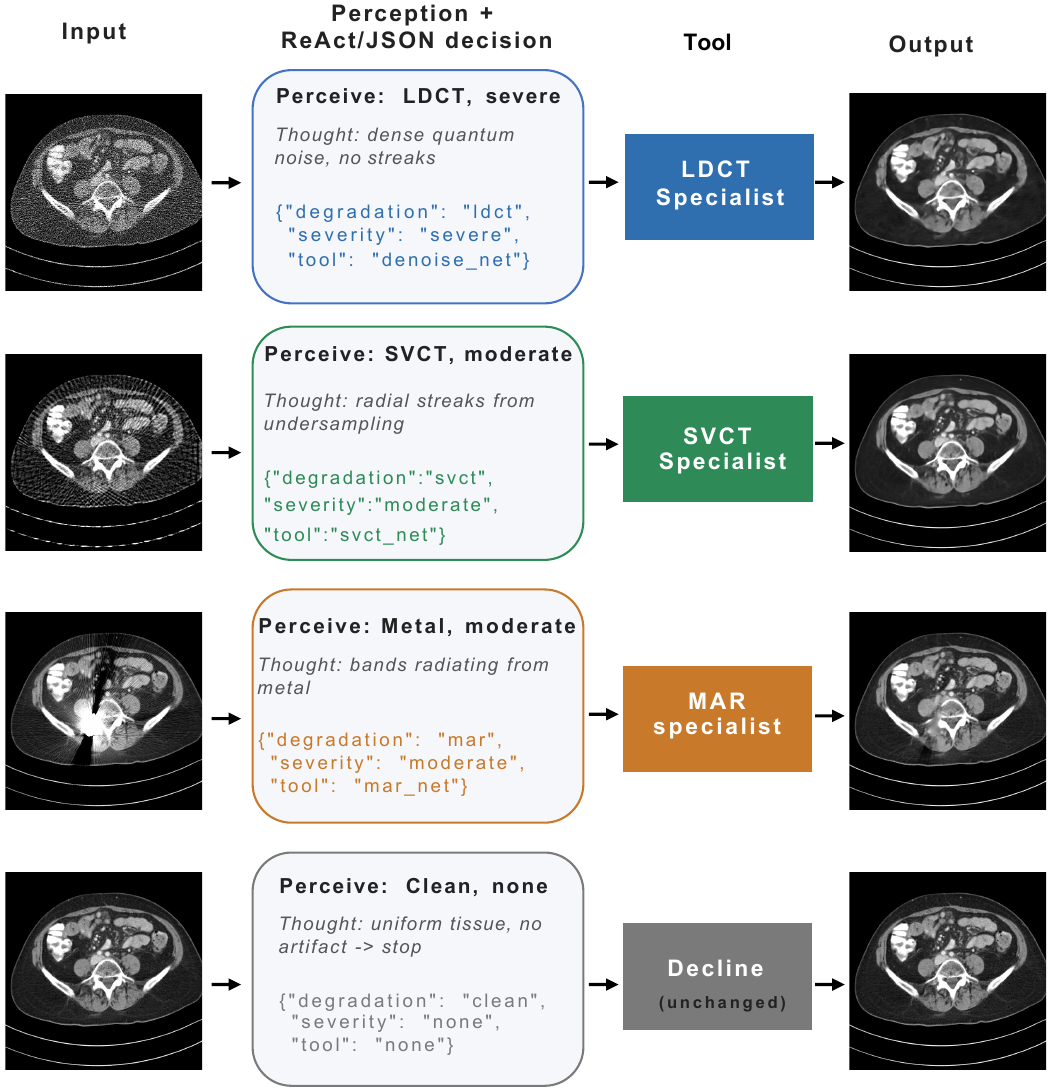}
\caption{End-to-end restoration with AgentCT. The controller perceives the degradation,
generates a structured routing action, invokes the corresponding specialist, and re-perceives
the restored result for verification.}
\label{fig:agent}
\end{figure}

\subsection{Examples of controller-guided imaging}
\label{sec:e2e}
To demonstrate restoration without manual model selection, we ran the complete
perceive-route-restore-verify process on representative cases (Fig.~\ref{fig:agent}). The
controller first analyzes the input, generates a structured routing action according to the
perceived degradation, invokes the corresponding tool, and returns the restored image for
verification; for a clean input, it outputs \texttt{none} and bypasses restoration. This
closed-loop behavior distinguishes AgentCT from conventional multi-model pipelines that require
the degradation type or restoration model to be specified manually, while each AgentCT specialist
remains independently designed for its degradation.

\subsection{Ablation: physics-aware architecture versus data separation}
\label{sec:ablation}
To test whether the improvement simply results from training separate networks per degradation,
we compared the dual-domain SVCT specialist with AMIR trained under both single-task and
all-in-one settings (Table~\ref{tab:ablation}). AMIR reached only 44.43~dB when trained on SVCT
alone and 43.57~dB under all-in-one training, whereas the dual-domain specialist achieved
46.04~dB with just 4.2~M parameters, compared with 23.5~M for AMIR. This indicates that the gain
does not come from separating the training data, but mainly from a task-specific architecture
that explicitly incorporates reconstruction-domain information through sinogram completion and
data consistency. The result supports the central motivation of AgentCT: to route different
degradations to heterogeneous, physics-aware specialists rather than to process all inputs with
a shared architecture.

\begin{table}[htbp]\centering
\caption{Effect of physics-aware specialization on SVCT (overall PSNR/SSIM over the three
sparsity levels).}
\label{tab:ablation}
\begin{tabular}{lc}
\toprule
Method & SVCT overall PSNR / SSIM \\
\midrule
AMIR single-task (SVCT only) & 44.43 / .973 \\
AMIR all-in-one (\{LDCT, SVCT, MAR\}) & 43.57 / .966 \\
\textbf{AgentCT (dual-domain, 4.2\,M)} & \textbf{46.04 / .982} \\
\bottomrule
\end{tabular}
\end{table}

\section{Discussion}
\label{sec:discussion}
Our results highlight three advantages of AgentCT. First, different CT degradations are handled
by specialists designed for their specific characteristics rather than forced into a single
shared architecture; this lets SVCT and MAR exploit reconstruction-domain physics and yields
better overall performance than the all-in-one baselines. Second, AgentCT demonstrates the
feasibility of the agentic paradigm in CT imaging by integrating perception, decision-making,
restoration, and verification into an adaptive, end-to-end workflow in which a single controller
both routes and explains its decision in natural language. Third, the framework is modular and
extensible: each AgentCT specialist can be replaced by a more advanced model, and new degradations and
tools can be added without redesigning the whole system, so universal CT imaging is achieved at
the system level through agentic orchestration rather than through one universal network.

Beyond aggregate fidelity, the physics-aware design also matters for the quantitative use of CT~\cite{denker2026stability}.
Preserving Hounsfield-unit accuracy is essential when restored images feed downstream tasks such
as attenuation-based tissue characterization or radiotherapy dose calculation, and enforcing
measurement consistency in the sinogram domain keeps the reconstruction faithful to the acquired
data rather than hallucinating structure. The low RMSE in Hounsfield units and the small residual
error observed in the error maps indicate that AgentCT preserves quantitative CT values rather
than merely improving perceptual similarity, which is the property most relevant to downstream
diagnostic and dosimetric use.

Several limitations remain. First, the present experiments consider one degradation at a time,
whereas real scans may contain multiple simultaneous degradations that require multi-step tool
planning. Second, the framework has not yet been validated on independently collected real
clinical degraded CT across different scanners, institutions, and acquisition protocols. Third,
AgentCT currently focuses on restoration with a predefined tool set; as multimodal language
models and medical agents mature~\cite{nath2025vilam3}, future systems could connect reconstruction with
segmentation, diagnosis, and clinical decision support, and reason jointly over images and other
clinical information~\cite{saab2026multimodal}. Addressing these points would move AgentCT from a restoration-oriented
agent toward a human-supervised intelligent imaging system.

\section{Conclusions}
\label{sec:conclusions}
We proposed AgentCT, a vision-language agent for universal CT imaging that unifies degradation
perception, tool routing, image restoration, and result verification in an adaptive framework. A
single VLM controller identifies the degradation type and severity and dispatches physics-aware
specialists for LDCT, SVCT, and MAR. The controller achieved 95.7\% type accuracy and 93.9\%
severity accuracy, and AgentCT reached an overall PSNR of 46.18~dB, outperforming the all-in-one
AMIR baseline by 1.25~dB and the CT foundation model TAMP by 1.89~dB. These results show that
universal CT restoration can be achieved at the system level through agentic orchestration of
specialized models rather than a single universal network, and provide an initial step toward
agentic CT imaging in which the system perceives the imaging condition, makes decisions, and
coordinates appropriate imaging tools.

\bibliographystyle{unsrtnat}
\bibliography{references}

\end{document}